\documentclass[letterpaper, 10 pt, conference]{ieeeconf}  % Comment this line out if you need a4paper

\IEEEoverridecommandlockouts                              % This command is only needed if 
\usepackage{graphics} % for pdf, bitmapped graphics files
\usepackage{epsfig} % for postscript graphics files
\usepackage{enumerate} 
\usepackage{subfigure}
\usepackage{graphicx}
\usepackage{float}
\usepackage{color}
\usepackage{amsmath}
\usepackage{amsfonts}
\usepackage{hyperref}
\usepackage{multirow}
\usepackage{stackengine}
\usepackage[export]{adjustbox}
\usepackage[inkscapelatex=false]{svg}
\usepackage{algorithm}
\usepackage{algpseudocode}

\newcommand{\argmin}{\mathop{\rm arg~min}\limits}

\makeatletter
\newcommand\fs@betterruled{%
  \def\@fs@cfont{\bfseries}\let\@fs@capt\floatc@ruled
  \def\@fs@pre{\vspace*{5pt}\hrule height.8pt depth0pt \kern2pt}%
  \def\@fs@post{\kern2pt\hrule\relax}%
  \def\@fs@mid{\kern2pt\hrule\kern2pt}%
  \let\@fs@iftopcapt\iftrue}
\floatstyle{betterruled}
\restylefloat{algorithm}
\makeatother

\title{\LARGE \bf
INF: Implicit Neural Fusion for LiDAR and Camera
}

\author{Shuyi Zhou$^{1,2}$, Shuxiang Xie$^{1,2}$, Ryoichi Ishikawa$^{1}$, Ken Sakurada$^{2}$, Masaki Onishi$^{2}$ and Takeshi Oishi$^{1}$% <-this % stops a space
\thanks{$^{1}$The authors are with The Institute of Industrial Science, The University of Tokyo, Japan. Emails:
        {\tt\small \{zhoushuyi495, shxxie, ishikawa, oishi\}@cvl.iis.u-tokyo.ac.jp}}%}% <-this % stops a space
\thanks{$^{2}$The authors are with The National Institute of Advanced Industrial Science and Technology (AIST), Tokyo, Japan. Emails: 
        {\tt\small \{k.sakurada, onishi-masaki\}@aist.go.jp}}%}%
}

\begin{document}

\maketitle
\thispagestyle{empty}
\pagestyle{empty}

%%%%%%%%%%%%%%%%%%%%%%%%%%%%%%%%%%%%%%%%%%%%%%%%%%%%%%%%%%%%%%%%%%%%%%%%%%%%%%%%
\begin{abstract}
Sensor fusion has become a popular topic in robotics. However, conventional fusion methods encounter many difficulties, such as data representation differences, sensor variations, and extrinsic calibration. For example, the calibration methods used for LiDAR-camera fusion often require manual operation and auxiliary calibration targets. Implicit neural representations (INRs) have been developed for 3D scenes, and the volume density distribution involved in an INR unifies the scene information obtained by different types of sensors. Therefore, we propose implicit neural fusion (INF) for LiDAR and camera. INF first trains a neural density field of the target scene using LiDAR frames. Then, a separate neural color field is trained using camera images and the trained neural density field. Along with the training process, INF both estimates LiDAR poses and optimizes extrinsic parameters. Our experiments demonstrate the high accuracy and stable performance of the proposed method. 

\end{abstract}

%%%%%%%%%%%%%%%%%%%%%%%%%%%%%%%%%%%%%%%%%%%%%%%%%%%%%%%%%%%%%%%%%%%%%%%%%%%%%%%%
\section{Introduction}
With the rapid advancement in sensor technology, integrating measurements from multimodal sensors has become increasingly significant in many applications. The outputs of multiple sensors can be combined to reduce the uncertainty, resolve the ambiguity, and increase the robustness of a system \cite{hackett1990multi}. 
LiDARs and cameras are commonly used sensors in robotics. 
A LiDAR can provide precise geometric information, whereas a camera can provide visual information about a scene. Combining LiDAR and camera data can help perform tasks such as simultaneous localization and mapping (SLAM) \cite{zhong2021survey}, object recognition, and 3D reconstruction.

Sensor calibration, which ensures that all sensors function correctly in a unified coordinate system, is a fundamental process during sensor fusion. However, many challenges hinder conventional calibration and fusion methods. 
In many cases, manual operation and auxiliary calibration targets are required. 
Popular solutions involve the use of single or multiple plane boards as targets \cite{verma2019automatic, zhao2020extrinsic, sim2016indirect, yang2012simple}. 
Some researchers have accomplished target-less LiDAR-camera calibration \cite{ishikawa2018lidar, koide2023general, schneider2017regnet, iyer2018calibnet}. 
However, the accuracy and robustness of these methods depend largely on scene complexity \cite{ishikawa2018lidar}, and pre-trained models or additional prior knowledge is sometimes required \cite{schneider2017regnet, iyer2018calibnet}. 
Therefore, automatic calibration between LiDARs and cameras remains challenging. 

\begin{figure}
\centering
\vspace{2mm}
% \includesvg[width=8.5cm]{pics/overview.svg}
\includegraphics[width=8.8cm]{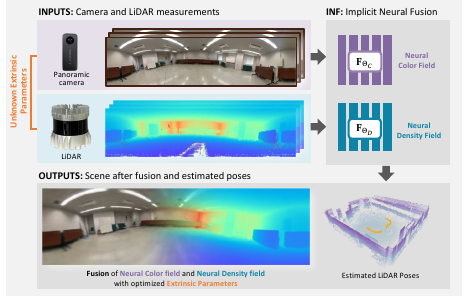}
\caption{Overview of INF. The proposed method takes camera and LiDAR observations as inputs to refine a neural color field and a neural density field. INF estimates the LiDAR pose of each frame and optimize the extrinsic parameters to further accomplish sensor fusion.}
\label{fig:overview}
\end{figure}

Following the trend of neural radiance fields (NeRF) \cite{mildenhall2020nerf}, we consider the use of implicit neural representations (INRs) to solve the above problems in sensor fusion. 
The volume density distribution given by an INR characterizes the geometric information of a space regardless of sensor type. 
Therefore, by obtaining a reliable volume density field, 
we may align all INRs using a unified volume density field and realize sensor fusion.  

In this paper, we propose implicit neural fusion (INF) for LiDAR and camera measurements. 
Fig.~\ref{fig:overview} shows an overview of the proposed method. 
We first use LiDAR frames to generate a neural density field because LiDARs provide reliable, high-quality geometric information. In this step, INF is able to estimate the LiDAR pose for each frame as well. 
Then, we combine camera images and the trained neural density field to obtain and further refine a color field. 
Meanwhile, we optimize extrinsic parameters for fusion between LiDAR and camera. 

Our contributions are summarized as follows:
\begin{itemize}
\item Provision of an implicit LiDAR-camera extrinsic parameter calibration method that does not require auxiliary calibration objects 
\item Introduction of a weighting technique for depth loss that improves the quality of neural density fields and the accuracy of LiDAR poses estimation
\item Proposal of a sequential LiDAR pose estimation method using neural volume density fields
\end{itemize}

\section{Related Work}
This section gives a brief review of studies on conventional LiDAR-camera fusion techniques, basic INR concepts, and camera pose estimation methods using INRs. 

\subsection{LiDAR-Camera Calibration} 
Based on the need for auxiliary targets, we classify LiDAR-camera calibration methods as target-based and target-less methods \cite{yan2022opencalib}. 

{\bf Target-based Methods: }
Target-based methods require a specified target in the scene. 
In most cases, the target can be a smooth plane with a feature pattern (such as checkerboard) \cite{verma2019automatic, zhang2004extrinsic, kim2019extrinsic, WANG2017Lidar_camera_cali} or markers \cite{li2020application, chai2018novel} or some solid colors \cite{chai2018novel, lee2017calibration}. 
The target can have different shapes, such as rectangle \cite{verma2019automatic, zhao2020extrinsic, zhang2004extrinsic,kim2019extrinsic, WANG2017Lidar_camera_cali, mishra2020experimental}, triangle \cite{ye2019extrinsic}, and sphere  \cite{lee2017calibration}. Applying multiple targets can improve calibration performance \cite{sim2016indirect, yang2012simple, geiger2012automatic}. 
The advantages of target-based methods are high stability and accuracy. 
However, preparing calibration targets is inconvenient and takes time and cost. There are no such objects in most public dataset and real-life application scenarios. 

{\bf Target-less Methods:} 
Target-less methods can be classified as appearance-, motion-, and geometry-based methods \cite{ishikawa2018lidar}: 
(1). {\it Appearance-based methods} detect and pair appearance features between a camera and a LiDAR. For instance, Pandey et al. \cite{pandey2012automatic} maximized the mutual information (MI) between image color and LiDAR reflectivity; Levinson and Thrun \cite{levinson2013automatic} and Yuan et al. \cite{ yuan2021pixel} detected and aligned the edges from both sensors' measurements; Schneider et al. \cite{schneider2017regnet} applied CNN to detect features from both of camera and LiDAR frames; Koide et al. \cite{koide2023general} utilized SuperGlue to match the features; \cite{sarlin2020superglue}  Zhu et al. \cite{zhu2020online} and Jiang et al. \cite{jiang2021calibrating} matched semantic information between the sensors.
(2). {\it Motion-based methods} use the hand-eye calibration approach. They first calculate the motion of camera and LiDAR separately. Then, the relative pose of the sensors can be obtained by comparing the trajectories of both sensors \cite{ishikawa2018lidar}. 
(3). {\it Geometry-based methods} extract 3D geometric information of the scene from multiple camera frames and directly find the correspondence in LiDAR scans \cite{chien2016visual}. CalibNet \cite{iyer2018calibnet} and CalibRCNN \cite{shi2020calibrcnn} used supervised neural networks to create depth map from images and compare it with LiDAR depths to align scenes. 

Appearance and geometry based methods require explicit 2D or 3D features; that is, they are highly dependent on the target scene.
Motion-based methods can hardly achieve high accuracy when the possible motions are restricted. 
While conventional works aim to match information from different domain data, our work uses a unified INR to align data derived from two sensors within the same domain.

\subsection{Implicit Neural Representations}
In addition to traditional approaches, such as point-based \cite{achlioptas2018learning, fan2017point}, mesh-based \cite{ranjan2018generating, wang2018pixel2mesh}, and voxel-based \cite{liao2018deep} methods,
researchers have recently proposed 
continuous, differentiable, implicitly defined networks to represent 3D scenes.
Park et al. and Mescheder et al. use signed distance functions or occupancy networks to describe surfaces implicitly \cite{park2019deepsdf, mescheder2019occupancy}. 
NeRF adopts volume rendering to encode appearance and features inside neural networks  \cite{mildenhall2020nerf}. 
%Take NeRF \cite{mildenhall2020nerf} as an example. 
It samples multiple 3D points according to the camera position; these sample points further pass through a fully connected network to obtain corresponding radiance and density values. \cite{mildenhall2020nerf}. 
Following certain works on NeRF, many pieces of research, such as \cite{yu2021pixelnerf, deng2022depth, barron2021mip}, propose several techniques that improve the performance of NeRF. 

However, to our best knowledge, few methods have been developed for integrating depth and color information obtained from independent sensors. 
Moreover, the NeRF and its variants usually require the camera poses as inputs.
These poses are often estimated using structure from motion (SfM) techniques; this additional process may harm the compactness of the whole system. 

\subsection{Camera Pose Estimation with INRs}
Researchers have attempted to solve camera pose estimation problems using INRs.
Many methods use the continuity of implicit neural networks to optimize input camera poses directly \cite{jeong2021self, lin2021barf, sucar2021imap}. 
Jeong et al. implements camera distortion models in their framework to optimize both camera intrinsic and extrinsic parameters simultaneously \cite{jeong2021self}. 
The implementation of pose parameters also varies. Jeong et al. applies a 6-vector representation for rotation \cite{jeong2021self}, while Lin et al. parameterized poses with the $\mathfrak{se}(3)$ Lie algebra \cite{lin2021barf}. Both methods manage to converge. 

Although the above mentioned methods demonstrate the feasibility of pose estimation, most existing methods perform targeting using RGB images. 
LiDAR depth is highly accurate and lightweight; 
thus, training speed and performance may be improved by the use of LiDAR data in pose estimation using INRs.

% % version 1: separate LiDAR and camera part
% \section{LIDAR POSE ESTIMATION WITH NEURAL DENSITY FIELD}
% \input{text/method_1.tex}
% \section{IMPLICIT CAMERA-LIDAR CALIBRATION}
% \input{text/method_2.tex}

% version 2: top down structure
\section{Implicit LiDAR-Camera Fusion}
\label{sec:overview}
% \begin{figure*}
%     \begin{center}
%     	\includegraphics[width=14.8cm]{pics/general1.png}
%     		\caption{We use LiDAR measurement data to train a neural density field which can represent the geometry of the space. Parts shown by dashed lines are optimized and updated. Notice that LiDAR poses indicated with red arrow, can also be estimated and optimized with proper implementation.}
%     	\label{fig:workflow1}
%     \end{center}
% \end{figure*}
% \begin{figure*}
%     \begin{center}
%     	\includegraphics[width=15.1cm]{pics/general2.png}
%     		\caption{We use camera data and trained density field to train a neural color field for color representation. Meanwhile, the extrinsic parameters can be optimized, which is the key part of sensor fusion. Parts shown by dashed lines are optimized and updated.}
%     	\label{fig:workflow1}
%     \end{center}
% \end{figure*}
\begin{figure*}
\centering
% \includesvg[width=17.7cm]{pics/general.svg}
\vspace*{0.2cm}
\includegraphics[width=17.7cm]{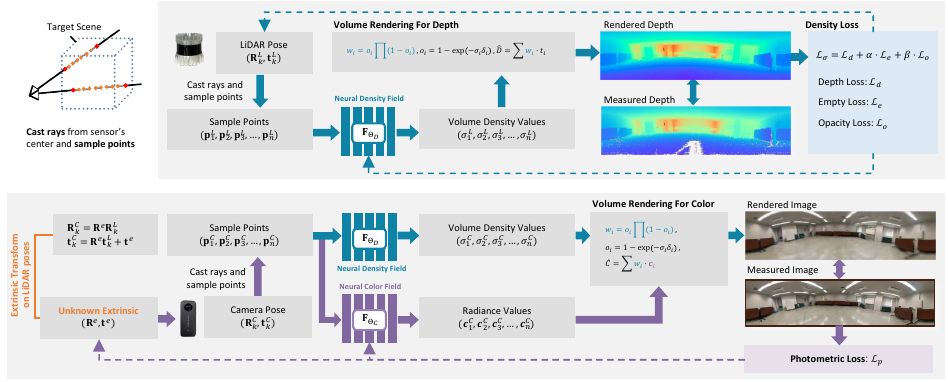}
\caption{General workflow of INF. We use LiDAR measurement data to train a neural density field that can represent the space geometry. Notice that LiDAR poses can also be estimated in this process. We also use camera data and trained density field to generate and refine a neural color field for color representation. Meanwhile, the extrinsic parameters can be optimized, which is the key part of sensor fusion. Processes shown by dashed lines indicate back-propagation, update and optimization.}
\label{fig:workflow1}
\end{figure*}

%In this study, 
We propose INF, an Implicit Neural Fusion system, which integrates LiDAR and camera data to estimate the extrinsic parameters of the sensors without calibration targets. 
Our method takes sequential LiDAR and camera measurements of a scene as inputs, and outputs the optimized extrinsic parameters, neural density field, and neural color field of the scene. 
In this section, we describe the general framework and workflow of the proposed fusion system and explain the details in procedures. 

\subsection{Assumptions}
We assume the use of a LiDAR and a camera combined with a rigid joint. 
Thus, the extrinsic parameters between the LiDAR and the camera should remain the same regardless of how the LiDAR-camera system moves. 
We also assume that each pair of camera frame and LiDAR frame is taken at the same time, in other words, temporally synchronized. 
The target scene must be static; that is, it should have no moving objects. 

\subsection{Problem Definition}
The LiDAR and camera at the same frame are capturing the scene from different locations and angles. The physical distance between the optical centers of a LiDAR and a camera makes scenes reconstructed by the two sensors misaligned. Therefore, we need to estimate the extrinsic parameters between the LiDAR and the camera to fuse the two reconstructed scenes together. 

We use LiDAR poses as bases. 
Although SfM techniques can powerfully reconstruct 3D scenes and estimate camera poses from image inputs, LiDAR point cloud alignment methods, such as the iterative closest point (ICP) algorithm still holds many advantages. First, LiDAR measurements can provide an absolute scale. Second, aligning LiDAR data takes lower computational costs. Third, LiDAR pose registration methods are more robust in textureless scenarios, especially in indoor cases. 
%Comparing with RGB images taken by cameras, scanning results from LiDAR can handle geometric features better with less noise. 
%As a result, 
Since the extrinsic parameters will not vary during the measurement, 
%due to the rigid setup. 
we may derive the camera poses using LiDAR poses and extrinsic parameters. 
Instead of optimizing every independent camera pose, optimizing only the extrinsic parameters makes the whole process easier. 

We first define the extrinsic parameters as $(\mathbf{R}^e,\mathbf{t}^e)$. We also let $\{\mathcal{P}^L\}$ be the set containing poses for all LiDAR frames. 
Define pose of $k$-th LiDAR frame pose as $\mathcal{P}_k^L = (\mathbf{R}_k^L,\mathbf{t}_k^L)$, and we can have $\mathcal{P}_k^L \in \{\mathcal{P}^L\}$. Then we may derive camera pose by applying $(\mathbf{R}^e,\mathbf{t}^e)$ on $\mathcal{P}_k^L$. Let $k$-th camera pose be $(\mathbf{R}_k^C,\mathbf{t}_k^C)$, we have 
\begin{equation}
\label{equ:camera-pose}
\mathbf{R}_k^C = \mathbf{R}^e \mathbf{R}_k^L,
\mathbf{t}_k^C = \mathbf{R}^e \mathbf{t}_k^L + \mathbf{t}^e. 
\end{equation}
Following the settings in \cite{lin2021barf}, all poses are parameterized with the $\mathfrak{se}(3)$ Lie algebra in this study. We further define two fully-connected neural networks as scene representations: {Neural Density Field} $\mathbf{F}_{\Theta_D} : \mathbf{x} \to \sigma$, and {Neural Color Field} $\mathbf{F}_{\Theta_C} : \mathbf{x} \to \mathbf{c}$. $\mathbf{F}_{\Theta_D}$ is used to represent the volume density distribution and $\mathbf{F}_{\Theta_C}$ is used to represent the color of every particles in the space.

%To conclude, 
Our INF system takes LiDAR frames $\{\mathcal{F}^L_1,\mathcal{F}^L_2 \cdots, \mathcal{F}^L_N\}$, camera frames $\{\mathcal{F}^C_1,\mathcal{F}^C_2 \cdots, \mathcal{F}^C_N\}$, and manually set initial value of  $(\mathbf{R}^e,\mathbf{t}^e)$ as inputs. By taking advantages of implicit neural representation, INF outputs LiDAR poses $\{\mathcal{P}^L\}$, optimized extrinsic parameters $(\mathbf{R}^e,\mathbf{t}^e)$, and refined neural networks $\mathbf{F}_{\Theta_D}$ and $\mathbf{F}_{\Theta_C}$. 
Note that, $(\mathbf{R}^e,\mathbf{t}^e)$ are applied to transform camera poses to world coordinates. %Their coordinate systems are combined with each other by the extrinsic parameters. 
Therefore, optimizing $(\mathbf{R}^e,\mathbf{t}^e)$ is equivalent to fusing $\mathbf{F}_{\Theta_D}$ and $\mathbf{F}_{\Theta_C}$. 

\subsection{Framework and Procedures}
\label{sec:workflow}
%$\mathbf{F}_{\Theta_D}$ defines the volume density, and $\mathbf{F}_{\Theta_C}$ defines the radiance distribution in the space. 
Fig.~\ref{fig:workflow1} gives an overview of the proposed INF system. 
Here, we explain the training procedures of the mentioned neural networks and parameters.
%maybe better to use subsubsection
\subsubsection{Training of Neural Density Field}
We train a neural density field $\mathbf{F}_{\Theta_D}$ using LiDAR measurement data. 
Notice that, LiDAR poses $\{\mathcal{P}^L\}$ can be obtained by two means. First, $\{\mathcal{P}^L\}$ is given by an Iterative Closest Point (ICP) method. Second, $\{\mathcal{P}^L\}$ is estimated together with $\mathbf{F}_{\Theta_D}$. The second method does not need output from pre-alignment so all procedures are implemented with INRs. 

Take $k$-th LiDAR frame $\mathcal{P}_k^L = (\mathbf{R}_k^L,\mathbf{t}_k^L)$ as an example.
Hereafter, subscript $k$ will be omitted for simplicity. 
First, we use $\mathcal{P}^L$ to calculate the laser rays in the world coordinates and then follow the techniques applied in \cite{mildenhall2020nerf} to sample several 3D points along the directions of laser rays. 
These samples are input into the defined $\mathbf{F}_{\Theta_D}$ after positional encoding. 
The corresponding volume density value for each sample is the output of $\mathbf{F}_{\Theta_D}$. 
Next, volume rendering is applied to obtain the estimated depth $\widehat{D}$. 
We use our designed loss functions to compare $\widehat{D}$ with observed depth $D$. 
Finally, we propagate the loss values back to optimize %neural network 
$\mathbf{F}_{\Theta_D}$. 
Notice that the loss functions will be the same if $\{\mathcal{P}^L\}$ is optimized here. 
The details of LiDAR pose estimation will be discussed in the following section. 

\subsubsection{Training of Neural Color Field}
We train a neural color field $\mathbf{F}_{\Theta_C}$ and optimize the extrinsic parameters simultaneously. Instead of directly defining camera poses, we use the derived camera poses as shown in equation \ref{equ:camera-pose}.
Then with the derived camera poses, we can obtain sample points along the camera rays. 
These sampled points are input to both $\mathbf{F}_{\Theta_D}$ and $\mathbf{F}_{\Theta_C}$ to get density values and radiance values. After volume rendering, estimated pixel color $\widehat{\mathbf{C}}$ is compared with observed color $\mathbf{C}$ to obtain photometric loss. 
We may give the initial value of $(\mathbf{R}^e,\mathbf{t}^e)$ manually before training. 
The most straightforward setting is {\it zero}, which means the camera is at the same pose and position of the LiDAR.  
This initial value setting may be inaccurate, leading to imperfect scene reconstruction and fusion. However, by back-propagating the loss function, we gradually optimize both $\mathbf{F}_{\Theta_C}$ and $(\mathbf{R}^e,\mathbf{t}^e)$.

\section{Implicit Neural Representations for Calibration and Scene Fusion}
% The MLP network used by NeRF and many following works is defined as $\mathbf{F}_\Theta: (\mathbf{x}, \mathbf{d}) \to (\mathbf{c}, \sigma)$ \cite{mildenhall2020nerf}, where $(\mathbf{x}, \mathbf{d})$ represent point position and viewing angle, and $(\mathbf{c}, \sigma)$ stands for estimated RGB color and volume density. These methods requires RGB image as supervision. Meanwhile, as indicated by some works, including depth can accelerate the convergence \cite{deng2022depth,sucar2021imap}.

%As introduced, the proposed INF includes two independent neural networks, $\mathbf{F}_{\Theta_C}$ and $\mathbf{F}_{\Theta_D}$. 
In this section, we describe color and depth rendering, the loss design for $\mathbf{F}_{\Theta_C}$ and $\mathbf{F}_{\Theta_D}$, and LiDAR poses estimation techniques. 
\subsection{Color and Depth Rendering}
As a core concept of NeRFs \cite{mildenhall2020nerf}, {\it volume rendering} has been extensively discussed. 
Besides RGB color rendering, volume rendering is widely applied to obtain estimated depths from volume density samples \cite{deng2022depth,sucar2021imap}. In this study, we use both color and depth rendering. 

Volume rendering is completed through sampling, querying, and integrating. We first sample 3D points along rays starting from the LiDAR/camera centers.
% a lot of-> More Specific amount
Then, we query the neural networks to obtain the corresponding properties, such as radiance and volume density. Let $\mathbf{p}_i = (x_i,y_i,z_i)$ be a 3D point, then we have the RGB color vector $\mathbf{c}_i = \mathbf{F}_{\Theta_C}(\mathbf{p}_i)$ and the volume density value $\sigma_i = \mathbf{F}_{\Theta_D}(\mathbf{p}_i)$. 
The rendered color and depth for each ray can be obtained by integrating the different properties. 
Color rendering follows $\widehat{\mathbf{C}} = \sum w_i\cdot \mathbf{c}_i$, whereas depth rendering follows $\widehat{D} = \sum w_i\cdot t_i$. 
$\mathbf{c}_i$ and $t_i$ represent the color and depth value, respectively, for each sample. 
$w_i$ denotes the transparency weight according to the setting in \cite{mildenhall2020nerf,sucar2021imap}, and it is calculated as $w_i = o_i \prod_{j = 1}^{i - 1}(1 - o_i)$, where $o_i = 1-\exp \left(-\sigma_i \delta_i\right)$. 
In this way, both color $\widehat{\mathbf{C}}$ and depth $\widehat{\mathbf{D}}$ are calculated. 

% \begin{figure}
% \centering
% \includesvg[width=8.5cm]{pics/total-loss.svg}
% \caption{Depth loss compares rendered depth and measured depth of each ray. Empty loss ensures there is no obstacles between LiDAR center and obejcts. Opacity loss describes whether the ray reaches objects or not. }
% \label{fig:density-loss-functions}
% \end{figure}

\subsection{Loss for Neural Color Field}
\label{sec:radiance-field}
After obtaining rendered RGB vector $\widehat{\mathbf{C}}$, we compare it with the observed color $\mathbf{C}$ to update  $\mathbf{F}_{\Theta_C}$. 
Given $n$ rays in total, the photometric loss $\mathcal{L}_{p}$ is obtained as, 
\begin{equation}
    \label{equ:photometric-loss}
        \mathcal{L}_{p} = \dfrac{1}{n}\sum\limits_{i=1}^n(\mathbf{C}_i - \widehat{\mathbf{C}}_i)^2.
    \end{equation}
$\mathbf{F}_{\Theta_C}$ and ($\mathbf{R}^e, \mathbf{t}^e$) are optimized simultaneously. The optimization process here is described as
\begin{equation}
\label{equ:extrinsic-color-opt}
\Theta_C, \mathbf{R}^e, \mathbf{t}^e = \argmin_{\Theta_C, \mathbf{R}^e, \mathbf{t}^e}  \mathcal{L}_{p}.
\end{equation}
\subsection{Loss for Neural Density Field}
\label{sec:density-field}
%The optimization for $\mathbf{F}_{\Theta_D}$ is more complicated. 
We design density loss as a combination of three parts: depth loss $\mathcal{L}_d$, empty loss $\mathcal{L}_e$, and opacity loss $\mathcal{L}_o$.
% The depth loss is calculated from rendered depth and observed depth.
% The empty loss and opacity loss are derived from the characteristics of LiDAR measurement.
The total loss for density field training is 
\begin{equation}
\label{equ:total-loss}
    \mathcal{L}_{\sigma} = \mathcal{L}_d + \alpha\cdot \mathcal{L}_e + \beta \cdot \mathcal{L}_o,
\end{equation}
where $\alpha$ and $\beta$ are used to scale the losses. 
% Figure \ref{fig:density-loss-functions} illustrates each part of $\mathcal{L}_{\sigma}$. 
The optimization process related to the neural density field is 
%please add argmin operation to indicate the optimized parameter
\begin{equation}
\label{equ:dense-pose-opt}
\Theta_D, \{\mathcal{P}^L\} = \argmin_{\Theta_D, \{\mathcal{P}^L\}}  \mathcal{L}_{\sigma}.
\end{equation}

{\bf Depth Loss: }Depth loss compares the ground truth depth $D_i$ and rendered depth $\widehat{D}_i$ as applied in many research \cite{deng2022depth, rematas2022urban}. 
Instead of directly using L1 norm as $|D_i - \widehat{D}_i|$ like in \cite{deng2022depth, rematas2022urban}, we use a weighting technique to emphasize the points near edges. 
Many LiDAR points are sampled at smooth planes, such as walls and grounds, which contribute too much to the loss. 
However, these points help little to the convergence of $\{\mathcal{P}^L\}$ because of the point-to-point nature of the error metric.
On the contrary, the points located around discontinuous parts, such as edges, are more significant. 
Therefore, we propose to weigh the depth loss according to whether the LiDAR ray reaches around the edges. 

The proposed weighting technique characterizes the extent of discontinuity for LiDAR points using the normal vector difference of neighboring points. 
Let $\eta_i$ be the depth weight of $i$-th ray, $\mathbf{p}_i$ be the point reached by $i$-th ray, and $\mathbf{n}_i$ be the normal vector with unit length at the location of $\mathbf{p}_i$. We label the rays according to their vertical and horizontal order in a scan. 
Let $(j,k)$ be the point index of $\mathbf{p}_i$ in the LiDAR frame. 
Then we define $\Phi_i$ as the set of normal vectors of neighboring points of $\mathbf{p}_i$. The neighboring points are defined based on their index, such as  $(j-1, k-1), (j-1, k)\dots (j+1, k+1)$. 
Thus $|\Phi_i| = 8$. 
The depth weight $\eta_i$ is calculated as, 
\begin{equation}
\label{equ:depth-weight}
    \eta_i = \dfrac{\lambda}{2}(1 - \dfrac{1}{|\Phi_i|}\sum_{\mathbf{n}_\phi \in \Phi_i}\langle\mathbf{n}_i, \mathbf{n}_{\phi}\rangle) + (1-\lambda).
\end{equation}
$\langle \cdot, \cdot \rangle$ is a vector dot product, $\lambda \in [0, 1]$ is a hyper-parameter representing how much edge points are emphasized. Larger $\lambda$ means the higher priority of edge points. Further, for all $n$ rays, we define the depth loss as, 
\begin{equation}
\label{equ:depth-loss}
\mathcal{L}_d = \dfrac{1}{\sum\limits_{i=1}^n \eta_i}\sum\limits_{i=1}^n \eta_i|D_i - \widehat{D}_i|.
\end{equation}

{\bf Empty Loss: }
Since $i$-th LiDAR ray reaches $\mathbf{p}_i$, then no obstacle should lie between the LiDAR center and $\mathbf{p}_i$. This means that the transparency weights should be $0$ for the sample points with depth $t < D_i$.
Consider $i$-th ray among all $n$ rays, where $K$ samples exist. Let the transparency weight of $k$-th sample point on $i$-th ray be $w_{(i,k)}$, and let the corresponding depth value be $t_{(i,k)}$. 
We define the $\lambda$-th sample point as the point that satisfies $t_{(i,\lambda)} < D_i - \epsilon< t_{(i,\lambda + 1)}$  on $i$-th ray. $\epsilon$ is a very small value modeling the error caused by discrete sampling. 
Then we may write the empty loss as, 
\begin{equation}
\label{equ:empty-loss}
\mathcal{L}_e = \dfrac{1}{n} \sum\limits_{i=1}^n \sum\limits_{k=1}^
% {\epsilon_i} 
{\lambda} (w_{(i,k)})^2 . 
\end{equation}

{\bf Opacity Loss: }
LiDAR can only reach points within the scanning range. For example, the scanning range of Ouster OS0 LiDAR is $0.25 \sim 35$m. If one laser does not detect any point within this range, the output depth value will be $0$. Opacity loss describes this phenomenon. 
We denote the opacity for $i$-th ray among $n$ rays as $y_i$. For the rays reaching objects, $y_i$ should be 1; for the rays that directly go outside the range, $y_i$ should be $0$. The opacity calculation is similar to the empty loss calculation, which is $\widehat{y}_i = \sum_{k = 1}^K w_{(i,k)}$, when there are $K$ sample points on $i$-th ray. We use binary entropy loss to characterize the opacity difference between rays, 
\begin{equation}
\label{equ:opacity-loss}
\mathcal{L}_o = -\frac{1}{N} \sum_{i=1}^N\left(y_i \cdot \log \widehat{y}_i+\left(1-y_i\right) \cdot \log \left(1-\widehat{y}_i\right)\right) .
\end{equation}

\begin{figure}
    \centering
    % \includesvg[width=8.5cm]{pics/dog-system.svg}
    \vspace*{0.1cm}
    \includegraphics[width=8.5cm]{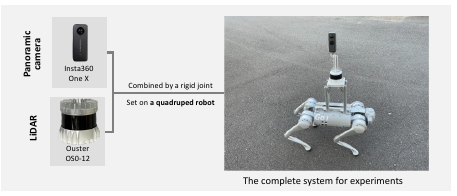}
    \caption{Experiment setup. We use a panoramic camera and a LiDAR. They are connected with a rigid joint and set on a quadruped robot.}
    \label{fig:dog}
\end{figure}

\subsection{LiDAR Pose Estimation}
\label{sec:lidar-pose}
As stated in Section \ref{sec:workflow}, instead of using LiDAR poses given by ICP methods, we can estimate $\{\mathcal{P}^L\}$ together with  $\mathbf{F}_{\Theta_D}$. In this way, we may increase the compactness of the system without losing accuracy in LiDAR pose estimation. 

Motivated by conventional SLAM systems, we apply the concept of local maps and keyframes to reduce accumulated errors. For LiDAR frames $\{\mathcal{F}^L_1,\mathcal{F}^L_2 \cdots, \mathcal{F}^L_N\}$, we define the keyframes as $\{\mathcal{F}^L_{\kappa_1},\mathcal{F}^L_{\kappa_2} \cdots, \mathcal{F}^L_{\kappa_n}\}$. 
Frames $\mathcal{F}^L_i$, where $ \kappa_a \leq i < \kappa_{a + 1}$, will form a local map. 
We set $\kappa_1 = 1$, and select keyframes based on the distance between the current frame and the previous keyframe. If the distance is larger than $\tau$, we will consider the current frame as a keyframe. 

We optimize within each local map to refine the density field and estimate relative poses between normal frames and the keyframe. The detailed process is shown in Alg. \ref{alg:lidar-pose}.

% In the $i^{\text{th}}$ local map, the density field $\mathbf{F}_{\Theta_D}$ is first trained by the keyframe $\mathcal{F}_{K_i}$. LiDAR frames $\mathcal{F}_{K_i+1}, \cdots,\mathcal{F}_{K_{i+1}}$ are then involved respectively into the computational graph to optimize their poses. After the poses are optimized one by one, we use all of the frames from $\mathcal{F}_{K_i}$ to $\mathcal{F}_{K_{i+1}}$ to optimize the density field $\mathbf{F}_{\Theta_D}$ and poses $\mathcal{P}_{K_i+1}^L, \cdots,\mathcal{P}_{K_{i+1}}^L$ together. 

\begin{algorithm}
\caption{LiDAR Poses Optimization}
\label{alg:lidar-pose}

\begin{algorithmic}[1]
% \Require $n \geq 0$
% \Ensure $y = x^n$

\Statex \textbf{Input: }$\{\mathcal{F}^L_1,\mathcal{F}^L_2 \cdots, \mathcal{F}^L_N\}$, $\mathbf{F}_{\Theta_D}$, $\tau$
\Statex \textbf{Output: } Optimized ${\Theta_D}$, $\{\mathcal{P}^L\}$
\State $i \gets 1$ \Comment{LiDAR frame index}
% \State $\mathcal{E} \gets \{\mathcal{F}_i^L\}$ \Comment{keyframe set}
\While {$i < N$}  
\State \textbf{Optimize} $\mathbf{F}_{\Theta_D}$ \textbf{with} $\mathcal{F}^L_i$ 
% \Comment{Base frame}
\State $d \gets 0$ \Comment{distance $d$ to previous keyframe}
\State $c \gets i$ \Comment{current frame index $c$}
    \While {$d < \tau$} 
        \State $c \gets c+1$
        \State \textbf{Estimate} $\mathcal{P}^L_c$ \textbf{with} $\mathbf{F}_{\Theta_D}$ \textbf{and}
$\mathcal{F}^L_c$
    \State $\mathcal{P}^L_{c+1} \gets \mathcal{P}^L_c$ 
    \Comment{set initial pose of next frame}
    \State $d \gets \|\mathcal{P}_c^L - \mathcal{P}_{c-1}^L\|_2$
    \EndWhile
    \State $\{\mathcal{F}_l\} \gets \{\mathcal{F}^L_i, ..., \mathcal{F}^L_{c-1}\}$ \Comment{local map}
    
    \State \textbf{Optimize} $\mathbf{F}_{\Theta_D}$ \textbf{and} $\{\mathcal{P}^L_{i+1},...,\mathcal{P}^L_{c-1}\}$ \textbf{with} $\{\mathcal{F}_l\}$
    \State $i \gets c-1$ \Comment{next keyframe}
    % \State $\mathcal{E} \gets \mathcal{E} \cup \{\mathcal{F}^L_i\}$
\EndWhile
% \State $\{\mathcal{P}^L_\kappa\} \gets \textbf{set of poses of frames in set } \mathcal{E}$
% \State \textbf{Optimize} $\mathbf{F}_{\Theta_D}$ \textbf{and} $\{\mathcal{P}^L_\kappa\}$ \textbf{with} $\mathcal{E}$

\end{algorithmic}
\end{algorithm}

\section{Experiments}
\subsection{Experimental Setup}
\label{sec:setup}
% In our experiments, we use Ouster OS0-128 LiDAR and Insta360 One X. 
In our experiments, we use Ouster OS0-128, whose resolution is $1024\times 128$, and Insta360 One X, whose resolution is $6080\times 3040$. 
We fix the camera on the LiDAR using a rigid joint on a Unitree Go1 quadruped robot as shown in Fig.~\ref{fig:dog}. 
When taking a shot, we move the quadruped robot by a small step and then capture a camera image and a LiDAR scan simultaneously after the robot stays still to ensure temporal synchronization.

\begin{figure*}
\centering
\vspace*{0.2cm}
% \includesvg[width=17.7cm]{pics/measured-rendered.svg}
\includegraphics[width=17.7cm]{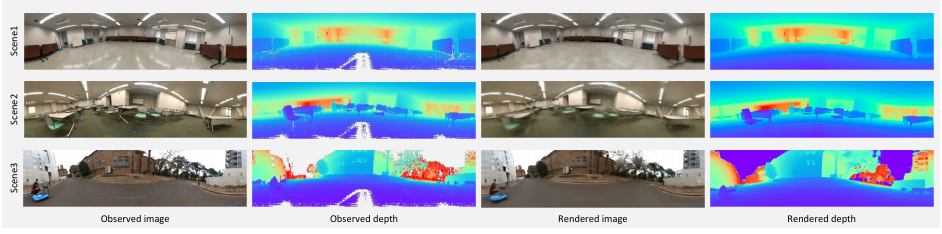}
\caption{Observed and rendered scenes. RGB images taken by panoramic camera and point clouds captured by LiDAR for all three scenes are shown on the left part. On the right side, we also show images rendered by the neural color field and depth rendered by the neural density field.}
\label{fig:scenes}
\end{figure*}

Fig.~\ref{fig:scenes} shows the three captured scenes. The first scene is a classroom with no occlusions; the second scene is a meeting room with various objects, such as desks and chairs; the third scene is an outdoor scene with trees and buildings. We use 30 frames for each scene in the following experiments. 
% We conduct the experiments 5 times and take the average of the results in the case of using INF to estimate LiDAR poses.

% \usepackage{multirow}
\begin{table*}
\caption{Results of Extrinsic Calibration using Different Methods}
    \centering
\begin{tabular}{|c|cccc|cccc|cccc|}

\hline
\multirow{3}{*}{}                       & \multicolumn{4}{c|}{Scene1: Classroom}                          & \multicolumn{4}{c|}{Scene2: Meeting Room}                       & \multicolumn{4}{c|}{Scene3: Outdoor}                            \\
                                        & \multicolumn{4}{c|}{Translation [m] \& Rotation [$^\circ$] Error} & \multicolumn{4}{c|}{Translation [m] \& Rotation [$^\circ$] Error} & \multicolumn{4}{c|}{Translation [m] \& Rotation [$^\circ$] Error} \\ \cline{2-13} 
                                        & x        & y        & \multicolumn{1}{c|}{z}        & r         & x         & y        & \multicolumn{1}{c|}{z}        & r        & x         & y         & \multicolumn{1}{c|}{z}        & r       \\ \hline

Mutual Info$\cite{pandey2012automatic}$ & 0.072    & 0.032    & \multicolumn{1}{c|}{0.259}    
&0.190 & 0.105 & 0.031 & \multicolumn{1}{c|}{0.155} &
% & 9.551     & 0.105 & 0.039 & \multicolumn{1}{c|}{0.154} & 
% 16.991   
16.902
&  0.055 & \textbf{0.015} & \multicolumn{1}{c|}{0.181} & 2.702 \\
Jump Edge$\cite{levinson2013automatic}$ & 0.019    & 0.031    & \multicolumn{1}{c|}{0.271}    & 3.288     & 
0.055 & 0.048 & \multicolumn{1}{c|}{0.171} &
% 0.055 & 0.056 & \multicolumn{1}{c|}{0.168} & 
% 15.987    
16.015
&  \textbf{0.001} & 0.078 & \multicolumn{1}{c|}{0.336} & 7.151 \\
Motion$\cite{ishikawa2018lidar}$        & 0.196    & 0.065    & \multicolumn{1}{c|}{1.450}    & 4.863    &
0.143 & 0.043 & \multicolumn{1}{c|}{0.728} &
 % 0.142 & 0.007 &  \multicolumn{1}{c|}{0.729} & 
% 7.652   
6.512
&  0.063 & 0.075 & \multicolumn{1}{c|}{0.783} & 0.868 \\
Continuous Edge$\cite{yuan2021pixel}$   & 0.619 & 0.332 & \multicolumn{1}{c|}{0.738} & 21.473  &  
% 0.085 & 0.105 & \multicolumn{1}{c|}{0.217} & 
% 17.962    
0.085 & 0.094 & \multicolumn{1}{c|}{0.222} &
17.690
&  0.546 & 3.090 & \multicolumn{1}{c|}{2.313} & 71.215\\
INF w/  g.t. pose (ours)  & 
\textbf{0.005}    & 0.006    & \multicolumn{1}{c|}{0.020}     & \textbf{0.299}  &   
\textbf{0.004} & 0.010 & \multicolumn{1}{c|}{\textbf{0.009}} & \textbf{0.385}
% \textbf{0.004} & 0.009 & \multicolumn{1}{c|}{\textbf{0.009}} & \textbf{2.670}
% \textbf{0.003}    & \textbf{0.004}    & \multicolumn{1}{c|}{\textbf{0.022}}     & 0.329  &   0.005 & \textbf{0.004} & \multicolumn{1}{c|}{0.014} & \textbf{0.367}
 & 0.013 & 0.017 & \multicolumn{1}{c|}{0.020} & \textbf{0.285} \\
INF w/o g.t. pose (ours)   
& 0.009    & \textbf{0.004}    & \multicolumn{1}{c|}{\textbf{0.015}}    & 0.680    &
0.009 & \textbf{0.002} & \multicolumn{1}{c|}{0.016} & 0.511
% 0.009 & \textbf{0.003} & \multicolumn{1}{c|}{0.016} & 2.818
% & 0.012    & \textbf{0.004}    & \multicolumn{1}{c|}{0.026}    & \textbf{0.270}    &\textbf{0.004} & 0.005 & \multicolumn{1}{c|}{\textbf{0.010}} & 0.369
& 0.015 & 0.037 & \multicolumn{1}{c|}{\textbf{0.010}} & 0.500 \\ \hline
\end{tabular}
    \label{tab:result}
\end{table*}

\begin{figure}
\centering
\includegraphics[width=8.5cm]{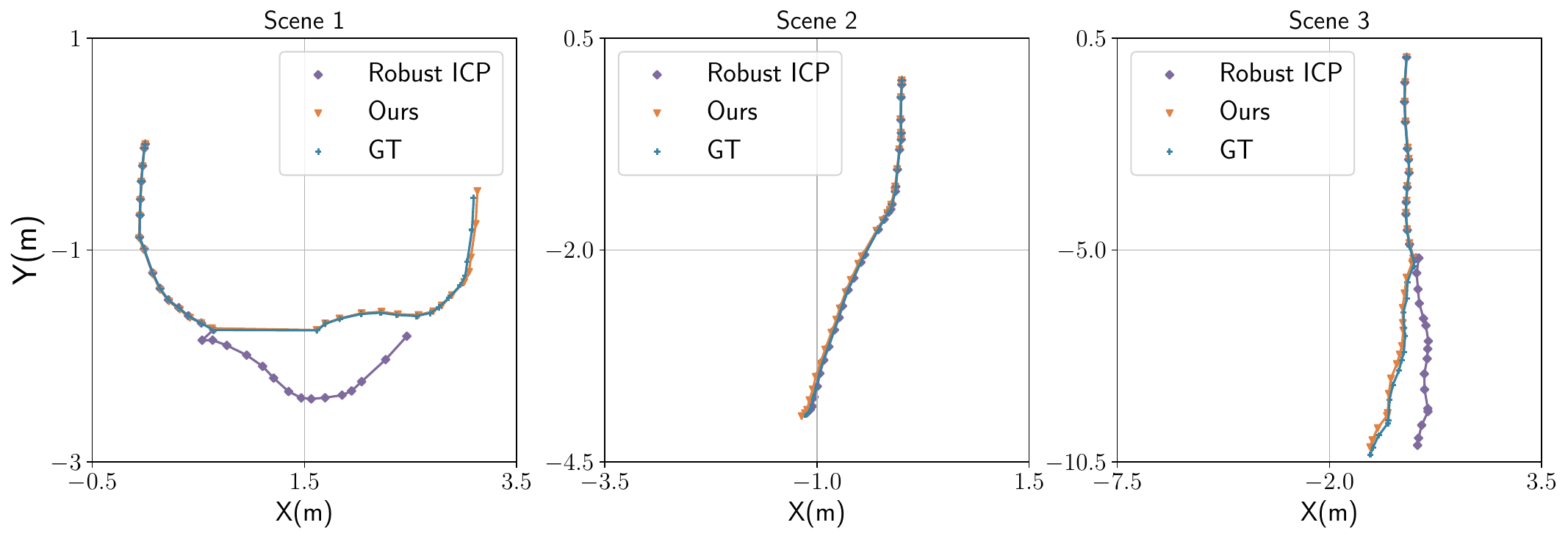}
\caption{
Comparison between estimated LiDAR poses using different methods. 
We compare the results obtained from INF (Ours), and traditional ICP methods without manual initialization.}
\label{fig:lidar-poses}
\end{figure}

\begin{table}[ht]
    \caption{Results of LiDAR Pose Estimation (APE \& RPE)}
    \centering

% \usepackage{multirow}
\begin{tabular}{|c|c|cc|}
\hline
& \multirow{3}{*}{Method} & \multicolumn{2}{c|}{APE / RPE} \\ \cline{3-4} 
&         & Translation [m]       & Rotation [$^\circ$]  \\  \hline
\multicolumn{1}{|c|}{\multirow{2}{*}{Scene1}} & Ours                    &   
\textbf{0.038 / 0.009} & \textbf{0.704 / 0.133}
% \textbf{0.018 / 0.007} & \textbf{0.401 / 0.188}
\\ \cline{2-4} 
\multicolumn{1}{|c|}{}                        & Robust ICP                  &    1.173 / 0.268 & 27.536 / 7.222   \\ \hline
\multicolumn{1}{|c|}{\multirow{2}{*}{Scene2}} & Ours                    &    
0.036 / 0.007 & 0.668 / 0.142  
% 0.042 / 0.007 & 0.315 / 0.214  
\\ \cline{2-4} 
\multicolumn{1}{|c|}{}                        & Robust ICP                  &   \textbf{0.017 / 0.002} & \textbf{0.099 / 0.029}      \\ \hline
\multicolumn{1}{|c|}{\multirow{2}{*}{Scene3}} & Ours                    &       \textbf{0.161 / 0.121} & \textbf{0.509 / 0.162} \\ \cline{2-4} 
\multicolumn{1}{|c|}{}                        & Robust ICP                  &        0.660 / 0.131 & 9.256 / 1.657  \\ \hline
\end{tabular}
    \label{tab:poses}
\end{table}

\begin{figure*}
\centering
% \includesvg[width=17.7cm]{pics/ablation.svg}
\vspace*{0.1cm}
\includegraphics[width=18.1cm]{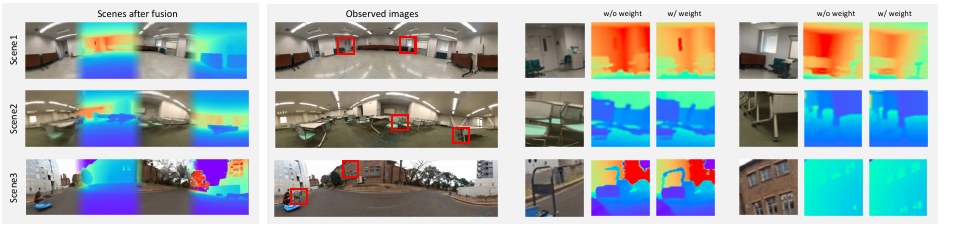}
\caption{{\bf Left: }Fusion of neural density field and neural color field. The two INRs are aligned. {\bf Right: }Qualitative evaluation of the rendered depth with or without weight introduced in section \ref{sec:density-field}. We can observe detailed structure more clearly when the weighting technique is involved.}
\label{fig:ablation}
\end{figure*}

\subsection{Ground Truth of Extrinsic Parameters}
%The ground truth LiDAR poses are calculated by manual initialization of frame poses and iterative optimization following a conventional alignment method \cite{oishi2005fast}. 

We obtain the ground-truth LiDAR-camera extrinsic parameters by minimizing the reprojection error $\mathcal{E}_r$ of the manually selected corresponding points between the camera frames $\mathbf{p}_i^C$ and LiDAR frames $\mathbf{p}^L_i$. The residual function is 
\begin{equation}
    \mathcal{E}_r = \sum_{i=1}^{N}
        T(\|\pi(\mathbf{R}^e\mathbf{p}^L_i + \mathbf{t}^e) - \mathbf{p}_i^C\|_2),
\end{equation}
where $\pi(\cdot)$ is the camera projection function and $N$ is the total number of selected points. $T(\cdot)$ is the Tukey loss function; it aims to avoid the error caused by manual selection. The reprojection errors of 90$\%$ of selected points are within 30 pixels.
% The accuracy for the ground-truth extrinsic parameters is about 1cm for translation and 0.2$^\circ$ for rotation.
% , which is defined as 
% \begin{equation}
%         T(r)= \begin{cases}\frac{c^2}{6}\left(1-\left[1-\left(\frac{r}{c}\right)^2\right]^3\right) & \text { if }|r| \leq c, \\ \frac{c^2}{6} & \text { otherwise. }\end{cases}
% \end{equation}

% In our experiment, we select 100 pairs of corresponding points for each dataset in total. And we repeatedly choose 50 pairs of points to calculate extrinsic parameters. 

\section{Results}
\subsection{Calibration Results}
The LiDAR-camera extrinsic parameters vary from scene to scene. However, we assume the camera and LiDAR are at the same relative pose throughout every single scene.
The initial values for $(\mathbf{R}^e,\mathbf{t}^e)$ are about 10$^\circ$/0.2m from ground truth values.  
The calibration results of different scenes are shown in Table \ref{tab:result}, where errors are calculated using absolute value. Fig.~\ref{fig:ablation} (left) also gives the visualization of scenes after fusion. 

Given the absence of specific target objects in our data, we use targetless methods based on appearance, motion, and geometry to perform comparative experiments. 
The compared methods are described below. 

\begin{enumerate}[(1).]
    \item {\bf Mutual Info: }The mutual information between LiDAR scan reflectivity and the camera image color is minimized \cite{pandey2012automatic}. 
    \item {\bf Jump Edge: }The edges in the LiDAR and camera frames are matched. Edges are detected by considering the discontinuity of LiDAR points \cite{levinson2013automatic}. 
    \item {\bf Motion: }This is a hand-eye calibration approach. The method uses ground-truth LiDAR poses and alternatively iterates the updating of extrinsic parameters and camera poses \cite{ishikawa2018lidar}. 
    \item {\bf Continuous Edge: }LiDAR point cloud edges are detected by cutting the point cloud into voxels and fitting the points into planes. Then, the edges in the LiDAR and camera frames are matched. This edge detection method requires a high-resolution LiDAR \cite{yuan2021pixel}. 
    \item {\bf INF w/ ICP LiDAR pose (ours): }Proposed INF system with the LiDAR poses given by ICP as inputs. 
    \item {\bf INF w/o ICP LiDAR pose (ours): }Proposed INF system without the LiDAR poses given by ICP. Instead, the LiDAR poses are estimated simultaneously with the density field.

    % \item {\bf ICP: }Conventional explicit points cloud alignment methods. We construct LiDAR points with raw input data from LiDAR, and obtain camera points using the multi-view stereo technique \cite{}. Aligning these two point clouds together will produce the result. 
    %TBC
\end{enumerate}

Generally speaking, our method has the highest accuracy among all methods.
Jump Edge \cite{levinson2013automatic} and Mutual Info \cite{pandey2012automatic} give small translation errors in the x and y axes, but these methods could not converge to the correct position in the z-axis and the rotation.
The proposed method is also robust enough to converge in the different scenes, whereas the other methods, such as Mutual Info and Jump Edge, fall into local minima easily. 
Continuous Edge does not function well due to the relatively low resolution of the LiDAR used. 
Moreover, the compared methods do not handle translation in the $z$ direction properly, whereas our method performs well in such a case. 
Fig.~\ref{fig:scenes} indicates the good rendering quality by INF of both the neural color and neural density fields. 
Fig.~\ref{fig:ablation} (left) shows that two INRs are aligned accurately. 

\subsection{LiDAR Poses Estimation Results}
We evaluate our estimated LiDAR poses using ground-truth poses. 
We also compare our method with an ICP method implemented by the public library Open3D \cite{Zhou2018, besl1992method, chen1992object}, which is denoted as Robust ICP. 
The results are shown in Table \ref{tab:poses}. Fig.~\ref{fig:lidar-poses} also gives a comparison between the LiDAR trajectory estimated by different methods and the ground truth. 
Results show that INF holds higher accuracy compared with Robust ICP in scene 1 and scene 3. 
Robust ICP cannot handle large displacement between frames, while INF maintains good performance. 
Robust ICP aligns point clouds directly, while our method aligns them with the trained neural density field; 
the accuracy of the neural density field affects the accuracy of pose estimation. 
In scene 2, though Robust ICP has a lower error, INF also shows reasonable accuracy.  

\subsection{Ablation Study on Depth Weight}
An ablation study is carried out for the weighting technique in Section \ref{sec:density-field} for LiDAR pose estimation and LiDAR-camera calibration. 
The qualitative result is shown in Table \ref{tab:weight}. We also evaluate the rendered results qualitatively as shown in Fig.~\ref{fig:ablation} (right). 
The performance difference with and without weighting indicates the effectiveness of our methods. 
In Fig.~\ref{fig:ablation} (right), we can also observe the improvement in scene reconstruction, especially in the detailed place with complex geometric features. 
Table \ref{tab:weight} indicates the reduction in pose error after applying the weighting technique in indoor scenes. 
In the outdoor case, the APE becomes larger with the weighting technique due to the existence of noises such as trees. 
However, the error is still small considering the total moving distance. Therefore, we may consider the proposed weighting technique helps the system with accuracy. 

\subsection{Ablation Study on Initial Value Setting}
\begin{figure}
    \centering
    \includegraphics[width=8.6cm]{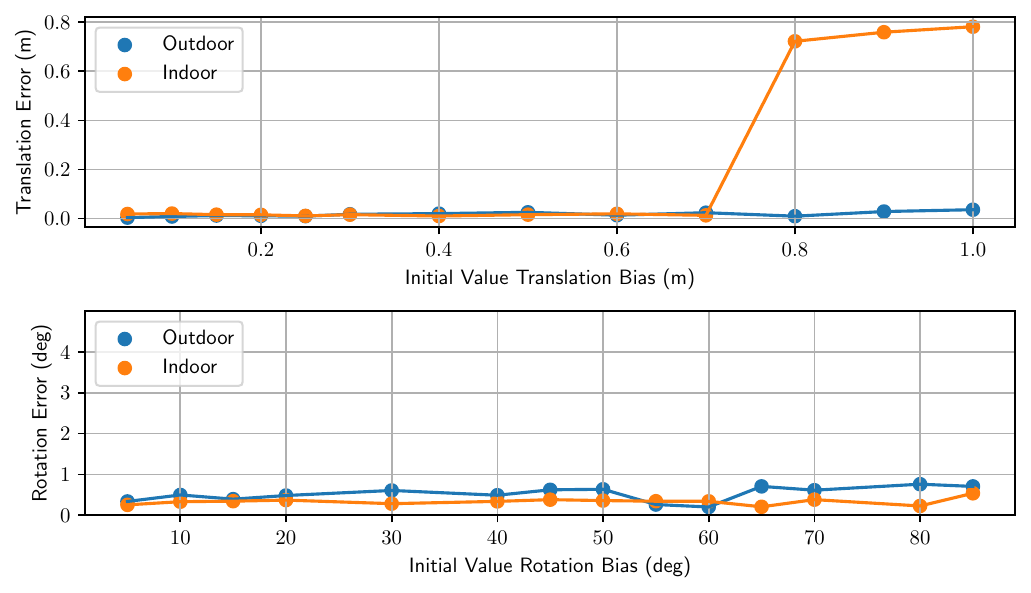}
    \caption{Calibration error with respect to initial extrinsic parameters bias. INF manages to converge even initial extrinsic parameters are set relatively far from ground truth.}
    \label{fig:conv}
\end{figure}

We also test the convergence of our method with respect to different initial values, including initial rotation values and initial translation values in both indoor and outdoor scenes. 
The result is shown in Fig.~\ref{fig:conv}. 
In our experiment, the convergence speed becomes a problem when a large initial bias is applied.
However, our method can converge even with an initial translation of $0.7$m or an initial rotation of $80^\circ$ away from the correct value. 
Compared with other neural network based methods, such as 1.5m/20$^\circ$ in RegNet \cite{schneider2017regnet}, 0.25m/7.5$^\circ$ in CalibNet \cite{iyer2018calibnet} and 0.25m/10$^\circ$ in CalibRCNN \cite{shi2020calibrcnn}, INF shows relatively high robustness against different initial value settings.

\begin{table}
    \caption{Results of Ablation Study for Weighting}
    \centering

%     \begin{tabular}{c|ccc}\hline
%         &Scene 1 & Scene2 & Scene3\\\hline
%     Depth loss with weighted& & & \\\hline
%     Depth loss without weighted&&&\\\hline
%     APE with weighted&0.038&0.036& 0.161\\\hline
%     APE without weighted&0.109& 0.054& 0.133\\\hline
%     RPE with weighted&0.009&0.007&0.121\\\hline
%     RPE without weighted&0.013&0.008&0.119\\\hline
%     \end{tabular}
%     \label{tab:weight}
% \end{table}

% \begin{tabular}{|c|c|cc|cc|}
% \hline
% \multicolumn{1}{|c}{}&\multirow{3}{*}{} & \multicolumn{2}{c}{APE of
% LiDAR Poses} & \multicolumn{2}{|c|}{Calibration Result} \\ \cline{3-6} 
% \multicolumn{1}{|c}{}&         & Translation      & Rotation     &  Translation & Rotation \\ 

% \multicolumn{1}{|c}{}&         &    [m]         &[$^\circ$]       & [m]     &[$^\circ$]        \\ \hline
% \multicolumn{1}{|c|}{\multirow{2}{*}{Scene1}} & w/o & 0.109  & 0.883 &0.030 & 0.733   \\ \cline{2-6} 
% \multicolumn{1}{|c|}{}                        & w/ & \textbf{0.038} & \textbf{0.704} & \textbf{0.018} & \textbf{0.647} \\ \hline
% \multicolumn{1}{|c|}{\multirow{2}{*}{Scene2}} & w/o & 0.054 & 0.764  & 0.020&\textbf{2.615} \\ \cline{2-6} 
% \multicolumn{1}{|c|}{}                        & w/  & \textbf{0.036 } & \textbf{0.668 } &  \textbf{0.018} & 2.804\\ \hline
% \multicolumn{1}{|c|}{\multirow{2}{*}{Scene3}} & w/o & \textbf{0.133 } & \textbf{0.371 }  &0.047 & \textbf{0.348} \\ \cline{2-6} 
% \multicolumn{1}{|c|}{}                        & w/ & 0.161    & 0.509   & \textbf{0.041} & 0.500   \\ \hline
\begin{tabular}{|cc|c|c|}
\hline
&&\multicolumn{2}{c|}{Error of Translation [m] / Rotation [$^\circ$]}\\\cline{3-4} 
\multicolumn{2}{|c|}{}& APE of LiDAR Pose Estimation & Calibration Error \\\hline
\multicolumn{1}{|c|}{\multirow{2}{*}{Scene1}}   & w/ & 
% \textbf{0.018} / \textbf{0.401} & \textbf{0.028} / \textbf{0.270}
\textbf{0.038} / \textbf{0.704} & \textbf{0.018} / \textbf{0.647}
\\ \cline{2-4} 
\multicolumn{1}{|c|}{} & w/o &
% 0.031 / 0.530 &0.029 / 0.418
0.109 / 0.883 &0.030 / 0.733  
\\ \hline
\multicolumn{1}{|c|}{\multirow{2}{*}{Scene2}} & w/  & 
% \textbf{0.042} / \textbf{0.315} &  \textbf{0.012} / 0.369
\textbf{0.036} / \textbf{0.668} &  
% \textbf{0.018} / 2.804
\textbf{0.018} / 0.511
\\ \cline{2-4} 
\multicolumn{1}{|c|}{} & w/o & 
0.054 / 0.764  & 
0.020 / \textbf{0.431} 
% 0.020 / \textbf{2.615} 
% 0.043 / 0.369  & 0.013 / \textbf{0.327} 
\\ \hline
\multicolumn{1}{|c|}{\multirow{2}{*}{Scene3}}                        & w/ & 0.161 / 0.509   & \textbf{0.041} / 0.500   \\ \cline{2-4} 
\multicolumn{1}{|c|}{} & w/o &\textbf{0.133} / \textbf{0.371}&0.047 / \textbf{0.348} \\ \hline

\end{tabular}
    \label{tab:weight}
\end{table}

\section{Conclusion}
In this paper, we propose INF for LiDAR-camera extrinsic calibration without auxiliary calibration targets. To our best knowledge, this work is the first to attempt to solve this problem using INRs. The proposed system exhibits good performance and robustness on several real-world datasets, and the proposed techniques are proven effective. Furthermore, unlike some target-less calibration and fusion methods, INF does not require prior knowledge or pre-trained models. Thus, our method could suit various scenarios. 

Future work will aim to improve the convergence speed of INF. We will closely examine and try to make good use of the geometric features hidden inside neural representations. We also plan to implement temporal synchronization in the system to enrich the input types of data. In addition, we aim to extend INF to other types of sensors, such as radar, temperature sensors, and event cameras. The sensor fusion process shall become more compact and convenient by leveraging the implicitness of the system.

\section*{Acknowledgement}
This work was partially supported by JST, PRESTO Grant Number JPMJPR22C4, Japan.


%%%%%%%%%%%%%%%%%%%%%%%%%%%%%%%%%%%%%%%%%%%%%%%%%%%%%%%%%%%%%%%%%%%%%%%%%%%%%%%%

\bibliographystyle{IEEEtran}
\bibliography{myrefs}

\end{document}